\documentclass[3p,times]{elsarticle}

\usepackage{ecrc}

\volume{00}

\firstpage{1}

\journalname{Robotics and Autonomous Systems}

\runauth{Elchanan Zwecher et al.}

\jid{raas}

\CopyrightLine{2022}{Published by Elsevier Ltd.}

\usepackage{amsmath} 
\usepackage{amssymb}  

\usepackage{graphicx}[pdftex]
\graphicspath{{figures/}}
\DeclareGraphicsExtensions{.pdf,.jpeg,.png}
\usepackage{multirow}
\usepackage{hhline}
\usepackage{makecell}
\usepackage{xcolor}

\usepackage[figuresright]{rotating}

\newcommand{\figref}[1]{\figurename~\ref{#1}}
\newcommand{\tabref}[1]{\tablename~\ref{#1}}

\newcolumntype{?}{!{\vrule width 1pt}}
\newcommand{\size}[2]{{\fontsize{#1}{0}\selectfont#2}}

\begin{document}

\begin{frontmatter}




\title{Deep-Learning-Aided Path Planning and Map Construction for Expediting Indoor Mapping}

\author{Elchanan Zwecher\fnref{heb}}
\author{Eran Iceland\fnref{heb}}
\author{Shmuel Y. Hayoun\fnref{techae}}
\author{Ahavatya Revivo\fnref{biu}}
\author{Sean R. Levy\fnref{hit}}
\author{Ariel Barel\fnref{techcs}}

\fntext[heb]{Computer Science Department, Hebrew University of Jerusalem, Jerusalem, Israel}
\fntext[techae]{Aerospace Engineering Department, Technion - Israel Institute of Technology, Haifa, Israel}
\fntext[biu]{Computer Science Department, Bar-Ilan University, Ramat Gan, Israel}
\fntext[hit]{Electrical Engineering Department, Holon Institute of Technology , Holon, Israel}
\fntext[techcs]{Computer Science Department, Technion - Israel Institute of Technology, Haifa, Israel}


\begin{abstract}
The problem of autonomous indoor mapping is addressed. The goal is to minimize the time to achieve a predefined percentage of exposure with some desired level of certainty. The use of a pre-trained generative deep neural network, acting as a map predictor, in both the path planning and the map construction is proposed in order to expedite the mapping process. This method is examined in combination with several frontier-based path planners for two distinct floorplan datasets. Simulations are run for several configurations of the integrated map predictor, the results of which reveal that by utilizing the prediction a significant reduction in mapping time is possible. When the  prediction is integrated in both path planning and map construction processes it is shown that the mapping time may in some cases be cut by over 50\%.
\end{abstract}


\end{frontmatter}



\section{Introduction}

Autonomous indoor mapping continues to be a relevant and challenging problem. One of the main difficulties is that the environment in which the mapping agent is intended to navigate is a priori unknown. When considering practical constraints on its energetic capacity, it is desirable for the agent to determine a strategy that minimizes the mapping time. Unlike other optimally-solved path planning problems where the domain is fully observable from the start (e.g. shortest path problem), in this setting the agent is equipped with on-board sensors (e.g. LIDAR, acoustic, stereo vision, etc.) and accumulates partial and often noisy information only about its immediate surroundings. In this case, the derivation of an online, globally optimal strategy is impossible.


The mapping of an unknown environment is called exploration. During exploration target points are selected sequentially in the initially unknown environment such that the agent progresses from point to point until the entire environment is mapped \cite{topiwala2018frontier}. This problem is strongly related to the NP-hard watchman route problem \cite{chin1986optimum}, in which we seek the shortest path that a watchman should take to guard an entire area with obstacles, given a map of the area. The main difference between the two is that in the watchman route problem the solution can be computed offline based on the prior knowledge of the layout of the area, whereas in the exploration problem the solution is constructed online based only on partial knowledge of the environment's architecture.
Contrary to outdoor scenes, indoor environments are more structured and adhere to at least some minimal basic design conventions. e.g., minimal room size, practical door and window locations, etc. In some cases these priors might be even greater, such as a building in which one floor's structure is indicative of another's layout, or a neighborhood where houses share similar architecture. 

Consider the problem of indoor mapping by an agent where the objective is to minimize the time by which a required portion of the environment is mapped with a certain degree of accuracy. A naïve approach would be to construct the map, commonly represented by an occupancy grid \cite{elfes1989using}, based exclusively on the observations. This is a classic exploration process that can be realised using the well known frontier based method \cite{yamauchi1997frontier}, which assumes no prior knowledge about the mapped area and selects target waypoints in any greedy manner. However, this and any similar approaches fail to utilise the underlying architectonic attributes that may be generalized and learned. In that respect, indoor mapping can be viewed as a problem lying somewhere between pure exploration where no prior knowledge is available and the watchman route problem where the environment is completely known.
In order to incorporate prior knowledge, the path planning problem can either be solved in the framework of a Markov decision process (MDP) where the occupancy grid acts as the observed state, or using heuristic methods (such as the frontier based algorithm) under a belief space planning (BSP) paradigm. In the former case one might consider the use of reinforcement learning \cite{zwecher2022integrating} to handle the path planning. In the latter the agent’s trajectory is planned in a more traditional fashion, while maintaining a belief map over the occupancy grid that is regarded as the actual observed state. 

This paper focuses on the second approach and presents the proposed methodology of integrating a trained image completion network with both the path planning and map construction procedures in order to expedite indoor exploration. The issue of planning under partial observations but with prior knowledge about their statistics is tackled by maintaining a belief map of the floorplan, generated by this deep neural network. This allows the agent to shorten the mapping duration, as well as enabling it to make better-informed decisions. This can be exploited in two ways: First, it is not necessary to visit the entire building to map it and partial observations can be used to map the rest unobserved area. This procedure of course shortened the required path. Second, since mapping is a sequential decision process based on partial observations, replacing the partial observation by predictions, can optimize the process even more, by choosing better waypoints. 

\section{Related Work}

Exploration of unknown environments is a fundamental task for autonomous robots. This task is traditionally solved by sequentially selecting target points that greedily maximizes a specified gain function. According to the seminal paper of Yamauchi \cite{yamauchi1997frontier}, such a point should be selected somewhere on the frontier: the boundary between open space and uncharted territory (though one can think of selecting a target point not necessarily on the frontier \cite{gonzalez2002navigation}) that maximizes the gain function. The various solvers of the exploration task differ from each other in the way of choosing the gain function. While the gain function of Yamauchi considers only the length of the path form the current agent's location to the frontier, later works take into account also the potential of exposure of candidate target points \cite{gonzalez2002navigation,topiwala2018frontier,holz2010evaluating}.

In past works, mapping and exploration by predictions based on partial observations was achieved by extracting geometric features representing the regularities of the indoor environment, as presented in \cite{luperto2020exploration,pimentel2018information}. However, in recent years neural networks have outperformed many traditional methods in estimation, classification and prediction. Among other things, they show impressive results in the realm of data generation and image restoration, also known as inpainting \cite{mao2016image,liu2018image,yu2019free}. They have also been used to recover partial 2D maps \cite{asrafexperience, pronobis2017learning, qi2020learning,caley2019deep, katyal2019uncertainty, shrestha2019learned}. However, most of the work on image restoration is concerned with denoising or "natural-looking" data generation, in the sense that since the exact missing or blurred observation cannot be obtained, a replacement to the unknown image that is as "natural" as possible is sought \cite{iizuka2017globally, yu2018generative}. Regarding 2D map completion, most previous work is concentrated on semantic classification \cite{pronobis2017learning, qi2020learning} and not in a pixel-by-pixel mapping. Moreover, the entire image is invariably used in to generate a local missing patch \cite{li2019houseexpo,shrestha2019learned,pimentel2018information}.

The approach in \cite{aydemir2012can} is based on graph theory, where each graph node represents a room, and each arc represents a connection between rooms, if it exists. Each room is given a 2D layout representation including centroid coordinates, a set of doorways that indicate a traversable path between this room and others if such exist, and a category label. Given an incomplete floorplan graph, the most likely next room category together with where it is connected to in the current graph or the most likely new path between the two rooms is predicted. One similarity between this work and ours is the creation of a new dataset, where their dataset was based on KTH campus data. 

Other works on navigation over 2D maps, look for an optimal local information gain action \cite{bai2017toward}, local matching topologies \cite{saroya2020online},  navigating using reinforcement learning techniques \cite{tai2016robot, niroui2019deep}, or exploiting existing plans of the building \cite{luperto2019robot}.

In \cite{shrestha2019learned} a map-predicting deep neural network was used to improve the estimated gain of candidate target points. The exposure potential of each point on the frontier was calculated based on the map predicted by the network. This work differs from ours in three aspects: First, the network in \cite{shrestha2019learned} is a variational autoencoder which is affected by random seeds and is designed to produce more of a natural looking output than an exact restoration, whereas our work includes a simple autoencoder with no random components which we believe to be better suited for generating precise maps. Second, in \cite{shrestha2019learned} only a small portion of the map is used to estimate missing regions, while our network does so with the entire map. Meaning, our network tries not only to capture local geometric features in the vicinity of the frontier, but the whole architectural plan of the building. Third and the most importantly, in \cite{shrestha2019learned}, the networks is used only for path planning, while the map is constructed exclusively by direct observations. Our work, on the over hand, proposes to rely on the output of the network for the map construction as well, allowing the agent to observe only a portion of the environment in order to complete its mission. 
While both works are done in the framework of frontier based exploration that incorporates statistics knowledge, the above choices significantly affect the network architecture as well as the evaluation of its performance.





\section{Problem Formulation}
Consider the problem of mapping a bounded indoor environment by an autonomous agent equipped with range sensors. The environment topology is assumed to possess some features common with an available dataset of floorplans. The problem is simplified by assuming that the localization problem is solved and neglecting any measurement noise. The desired constructed map, the boundary of which is known a priori, is given in the form of a 2D binary occupancy grid in which each cell may be either vacant, occupied or unknown. 
\begin{equation}
\label{eq:thresholds}
m(x,y)=\begin{cases}
c_{\text{free}}, & \text{if $(x,y)$ is observed free space}\\
c_{\text{obstacle}}, & \text{if $(x,y)$ is observed obstacle}\\
c_{\text{unknown}}, & \text{if $(x,y)$ is unobserved}\\
c_{\text{agent}}, & \text{the agent's location}.
\end{cases}
\end{equation}
The agent can move in any one of eight directions on the occupancy grid
\begin{equation}
\label{eq:actions_space}
\mathcal{A}\in\{\mathcal{\uparrow,\nearrow,\rightarrow,\searrow,\downarrow,\swarrow, \leftarrow,\nwarrow}\}.
\end{equation}
The objective is to minimize the time to reach a predefined percentage of coverage with some desired level of certainty. 

\section{Proposed Solution}
Any basic mapping system design includes a map constructor which is generally fed sensory inputs and a path planner which bases its decision making upon the existing constructed map. Normally, the constructed map simply includes the accumulated observations.

\begin{figure}[t]
  \includegraphics[width=\linewidth]{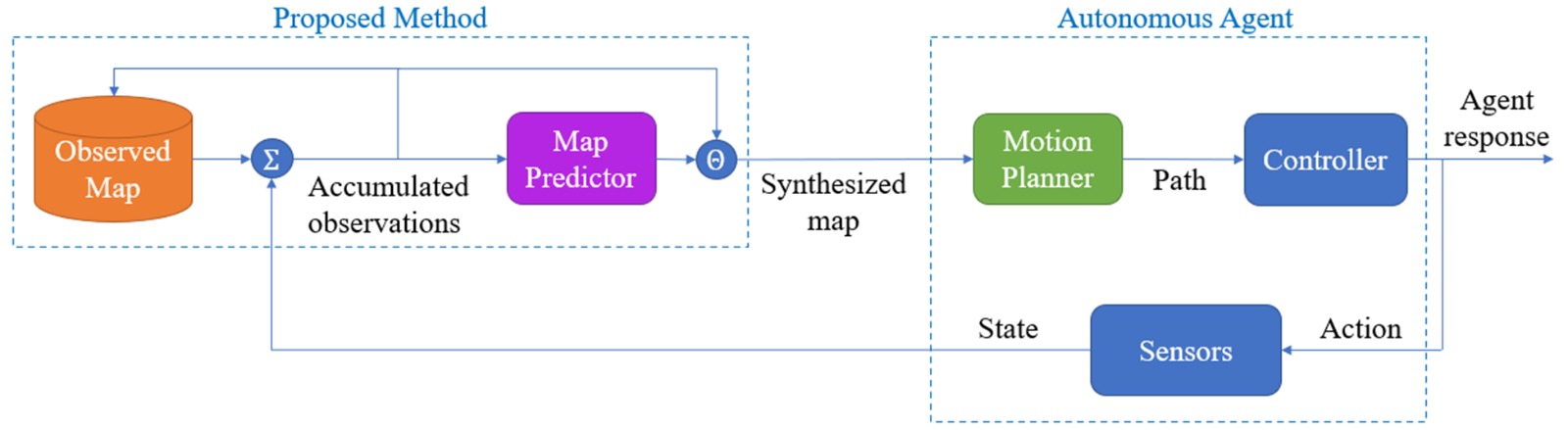}
  \caption{The proposed cascaded control scheme of the our agent. On the left hand side is our contribution including $\Sigma$ - the observations accumulation operator, and $\Theta$ - the observations and prediction synthesis operator. On the right hand side is a common control scheme of an autonomous agent.}
  \label{fig:block_diagram}
\end{figure}

The proposed solution consists of enhancing the performance of the path planning and map construction processes through the incorporation of learned experience. Concretely, the proffered mapping system includes a deep artificial image completing neural network, trained in a supervised manner on numerous examples of partial observation maps to output a prediction over the entire indoor environment. The mapping system in its entirety is depicted in Fig. \ref{fig:block_diagram}. The map constructor now receives the map prediction in addition to the accumulated observations, and produces a combined occupancy grid composed of the observations and the prediction in unobserved areas.

As discussed earlier, experience-based path planning has been proven to expedite the mapping process. Therefore, expected potential benefits include savings in observations (measurements) required for reaching the desired mapping goal as well as improved, more informed decision making on the path planner's part.

\section{Map Prediction}
The map predictor is a deep neural network tasked with recovering the entire structural map from given partial observations.
Essentially, the predictor is a function that takes a partially observed occupancy grid as an input and outputs a probability distribution over the entire grid,
\begin{equation}
    f_{\text{prediction}}:\mathcal{O} \rightarrow \left[0,1\right]^{h\times w},
\end{equation}
where $\mathcal{O}=\{c_{\text{free}},c_{\text{obstacle}},c_{\text{unknown}}\}^{h\times w}$ is the observation space. The prediction over each cell is the probability of it containing an obstacle, where the output value $0$ indicates certain free space and $1$ indicates certain obstacles. This probabilistic map is expected to be consistent with the given observations, i.e. exposed cells should be classified as they were observed with very high probability. In order to get a categorized probable map after obtaining a prediction from the network a thresholding is performed in the following manner over each cell in the occupancy grid:
\begin{equation}
\label{eq:thresholds}
f_{\text{threshold}}\left(p_m(x,y)\right)=\begin{cases}
c_{\text{free}}, & p_{m}(x,y) \leq \dfrac{1-\delta_{\text{free}}}{2}\\
\vspace{-8pt}\\
c_{\text{obstacle}}, & p_{m}(x,y) \geq \dfrac{1+\delta_{\text{obstacle}}}{2}\\
c_{\text{unknown}}, & \text{otherwise}.
\end{cases}
\end{equation}
where $p_m$ is the probability of wall and  $\delta_c \in \left[0, 1\right]$ is the threshold corresponding to category $c$ (can also be considered as the confidence level of category $c$).
Unlike other similar works in which both thresholds are trivially set to 0.5, the "unknown" category was added since it was perceived that in areas where very few observations exist obtaining a reliable distinct prediction is practically impossible. The probability of any cell, situated far away from any observations, being occupied will quite likely tend towards the general probability of any random cell being occupied, which is simply the ratio of occupied cells (walls) in the structure. This value is of course expected to be significantly less than 0.5. Therefore, the "unknown" category is intended to reflect when the prediction is unreliable. Generally, the thresholds determine the trade-off between the number of false positives and negatives and the map construction rate. The chosen set of values was obtained through trial and error until a good balance between the thresholded prediction's $F_1$-score and the mapping duration was found. Thus we obtain the map predictor 
\begin{equation}
    f_{\text{predictor}} = f_{\text{threshold}} \circ f_{\text{prediction}}: \mathcal{O} \rightarrow \mathcal{O}.
\end{equation}
The final prediction-based map is synthesized by overlaying the observed sections of the map.

\subsection{Architecture}
\label{sec:predictor_architecture}
\begin{figure}
  \includegraphics[width=\linewidth]{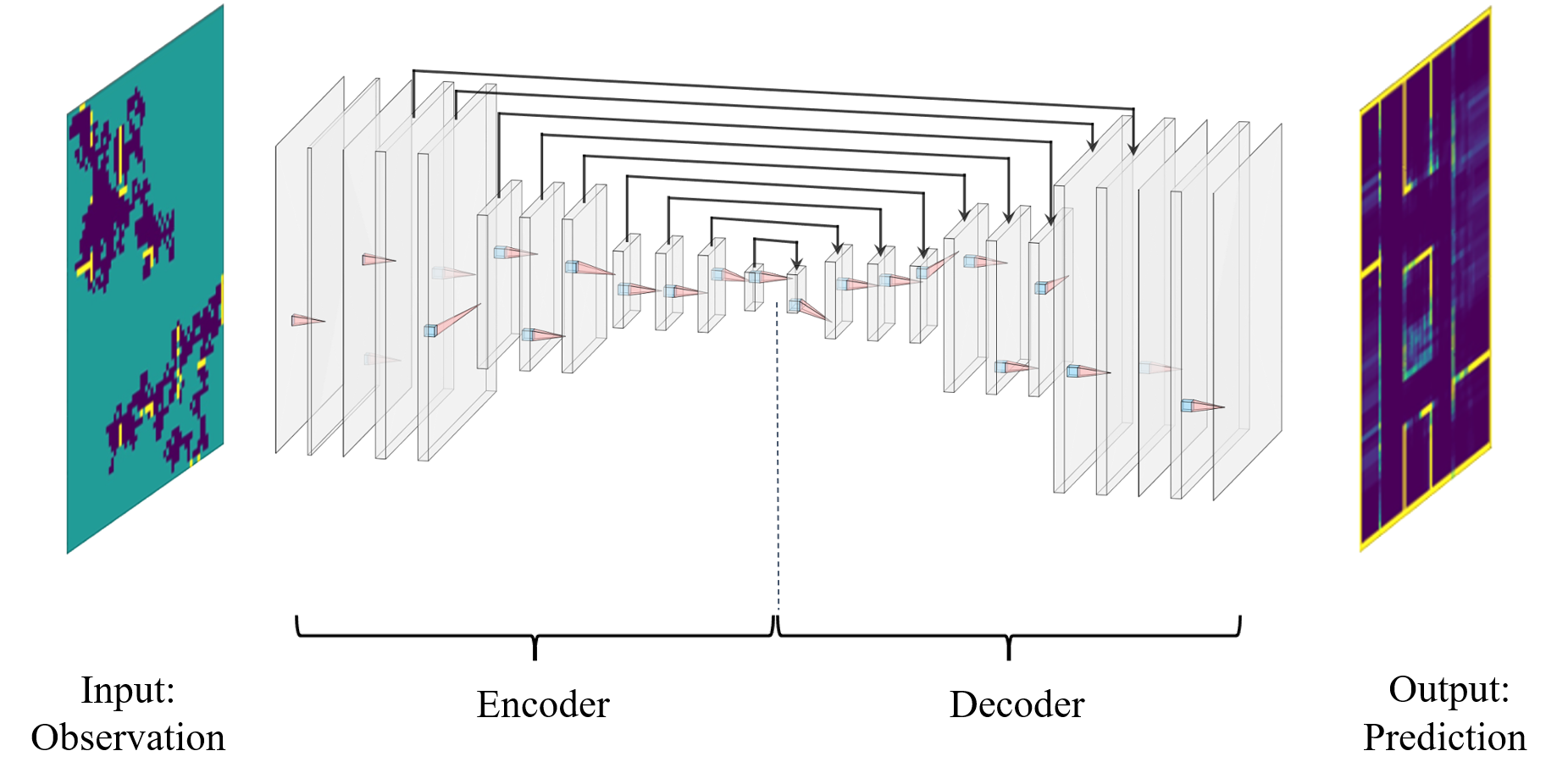}
  \caption{Map predictor architecture -- ResNet-styled convolutional autoencoder}
  \label{fig:nn_architecture}
\end{figure}

The map predictor, illustrated in Fig. \ref{fig:nn_architecture}, is a classifying network, implemented as a convolutional autoencoder with symmetric skip connections. As discussed above, image completion is frequently accomplished using data generation methods, such as generative adversarial networks (GANs) \cite{goodfellow2014generative} or variational autoencoders (VAEs)\cite{higgins2016early} to produce "natural-looking" images. However, since in the present case an exact reconstruction of the floorplan is coveted, simply generating "natural-looking" maps is insufficient. 

The network receives as input a three channel image where each channel corresponds to one of three categories: occupied, vacant or unknown. Each cell in the occupancy grid is represented by a triplet corresponding to these three categories, such that only one of them equals 1 while the two others are 0, also known as a one-hot encoding. The generated output is of the same dimensions as the input.

The core of the architecture is comprised of nine encoding layers and nine decoding layers. Each encoding layer contains 25 $3\times3$-sized kernels. Every third layer also includes a stride of 2 in order to obtain a reduced sized encoded representation of the input in the so-called latent space. Similarly, in an inverse manner, every third layer in the decoder incorporates 3D transposed convolution with a stride 2 in order to recover original map size.    
The addition of the symmetric skip-connection. which was reported by~\cite{mao2016image} to improve performance, is used not in order to enrich the hypothesis class (as done for more elaborate architectures), but mostly in order to improve the optimization process by aiding the back-propagation algorithm in overcoming the exploding/vanishing gradient problem \cite{he2015deep}. Moreover, in the present case there one might also see the sense in repeatedly "reminding" the subsequent layers in the network of the output from previous layers, since the observations should be regarded while estimating the unknown cells.

Two more dominant features were added to the network. The first was introducing an initial hidden layer tasked with augmenting the input so as to improve the optimization process. The added first layer is a convolution layer with a single 1$\times$1$\times$3 kernel without bias that takes the three-channeled input and outputs a single channel image, where each cell contains one of three learned optimal values representing the three possible categories. This new representation of the input enhances the network optimization during training. The second feature was stacking the input observation on to the input to the last hidden layer, such that the number of channels of that layer is increased by one. This operation, similar to the skip connection, serves as a "reminder" of the input immediately prior to the final map reconstruction stage. This is done with the intent of accentuating the observations in order to improve the quality of the reconstructed output.

\subsection{Optimization}
The neural net (before thresholding) is designed to output a probability map which the result of extensive training, during which the following optimization criteria is to be minimized:
\begin{equation}
    \mathcal{L} = -\sum_{i \in \mathcal{I}}\left[y_i\log(p_i) + (1-y_i)\log(1-p_i)\right].
\end{equation}
$y_i$ is the actual label of pixel $i$, i.e. $0$ if vacant and $1$ if occupied and, accordingly, $p_i$ is the probability of $i$ being occupied.
The chosen function $\mathcal{L}\left(\cdot\right)$, otherwise known as the loss function, is the binary cross entropy loss - a commonly used loss function for classification problems. This function measures the Kullback–Leibler divergence between the true distribution (i.e. label) and the predicted distribution over a pixel.

\subsection{Training Sets}
\begin{figure}[ht]
  \includegraphics[width=\linewidth]{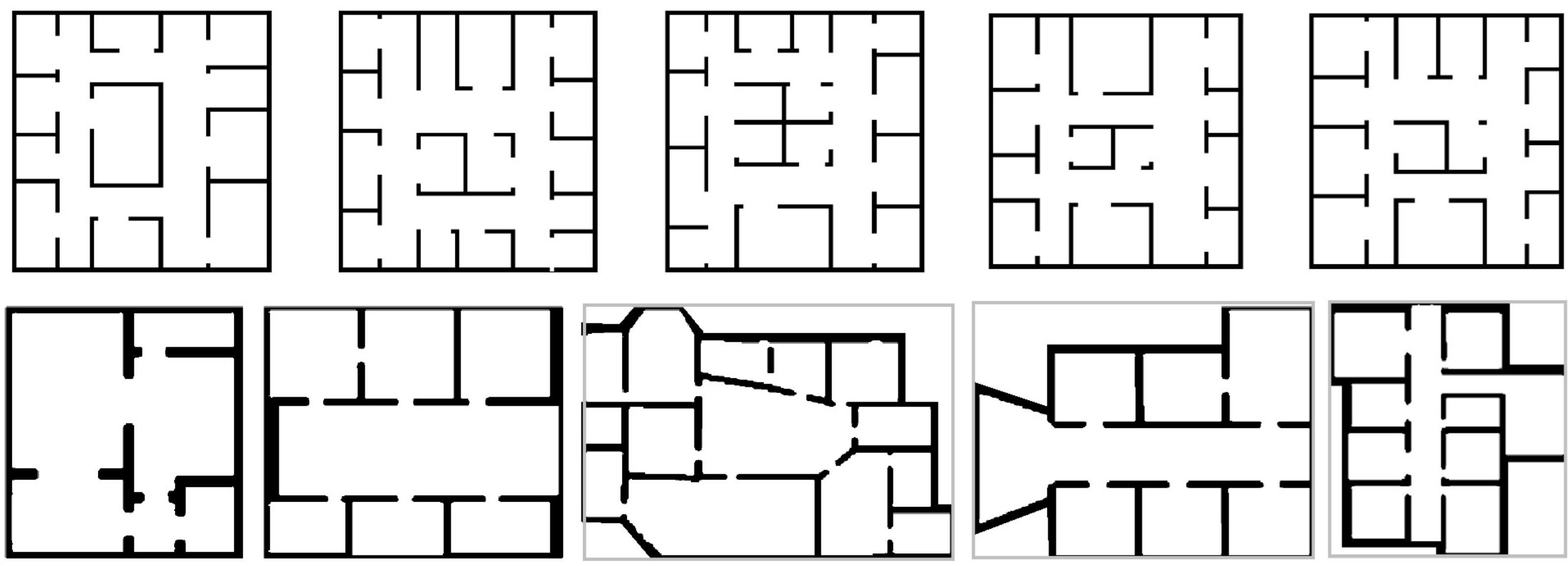}
  \caption{Illustrative floorplan examples. The top row includes examples from $\mathcal{D}_1$ and the bottom row includes examples from $\mathcal{D}_2$}
  \label{fig:dbs}
\end{figure}

\begin{figure}[htbp]
  \includegraphics[width=\linewidth]{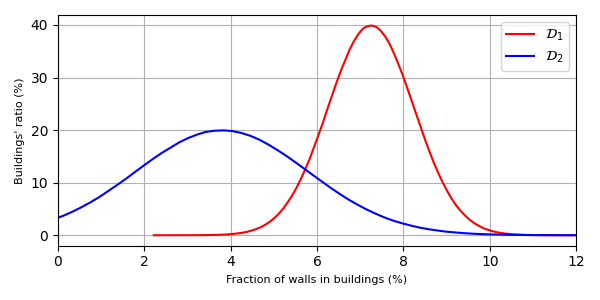}
  \caption{Fraction of walls distribution for the different datasets}
  \label{fig:db_statistics}
\end{figure}

The map predictor was trained on two distinct datasets, differing mainly in level of variance of the building layouts' geometrical features, as listed in \tabref{table:datasets}. Several examples from each dataset are displayed in \figref{fig:dbs}. \figref{fig:db_statistics} exhibits the distributions of the buildings' fraction of walls in each dataset, illustrating their structural differences.

\begin{table}[h]
\caption{Datasets characteristics}
\label{table:datasets}
\begin{center}
\begin{tabular}{clllll}
\cline{4-5}
\multicolumn{1}{l}{} & \multicolumn{1}{l}{} & \multicolumn{1}{l|}{} & \multicolumn{2}{c|}{\multirow{2}{*}{\textit{\textbf{\size{11}{Datasets}}}}} \\
\multicolumn{1}{l}{} & \multicolumn{1}{l}{} & \multicolumn{1}{l|}{} & \multicolumn{1}{l}{} & 
\multicolumn{1}{l|}{} \\
\hhline{~~~|==|} 
\multicolumn{1}{l}{} & \multicolumn{1}{l}{} & \multicolumn{1}{l|}{} & \multicolumn{1}{c|}{\multirow{2}{*}{$\mathcal{D}_1$}} & 
\multicolumn{1}{c|}{\multirow{2}{*}{$\mathcal{D}_2$}} \\
\multicolumn{1}{l}{} & \multicolumn{1}{l}{} & \multicolumn{1}{l|}{} & 
\multicolumn{1}{c|}{} & \multicolumn{1}{c|}{} \\
\hline 
\hhline{|~||~~|--} 
\hhline{|~||~~|--}
\multicolumn{1}{|c||}{\multirow{5}{*}{\rotatebox{90}{\parbox{3.15cm}{\textit{\textbf{\size{11}{Characteristics}}}}}}} & \multicolumn{2}{l|}{Source} & \multicolumn{1}{?l|}{\begin{tabular}[c]{@{}l@{}}Independently generated\end{tabular}} &
\multicolumn{1}{l?}{\begin{tabular}[c]{@{}l@{}}HouseExpo \cite{li2019houseexpo}\end{tabular}} \\ 
\cline{2-5} 
\multicolumn{1}{|c||}{} & \multicolumn{2}{l|}{\multirow{2}{*}{Size}} & \multicolumn{1}{?l|}{\multirow{2}{*}{50,000}} & 
\multicolumn{1}{l?}{\multirow{2}{*}{35,126}} \\ 
\multicolumn{1}{|c||}{} & \multicolumn{2}{l|}{} & \multicolumn{1}{?l|}{} & 
\multicolumn{1}{l?}{} \\ 
\cline{2-5} 
\multicolumn{1}{|c||}{} & \multicolumn{2}{l|}{Contour} & \multicolumn{1}{?l|}{\begin{tabular}[c]{@{}l@{}}Identical\\ Convex (rectangle)\end{tabular}} & 
\multicolumn{1}{l?}{\begin{tabular}[c]{@{}l@{}}Varying\\ Convex \& concave\end{tabular}} \\ 
\cline{2-5} 
\multicolumn{1}{|c||}{} & \multicolumn{1}{c|}{\multirow{6}{*}{\rotatebox[origin=c]{90}{\parbox[c]{1.2cm}{\centering Variance}}}} &
\multicolumn{1}{l|}{\multirow{2}{*}{Size}} & \multicolumn{1}{?l|}{\multirow{2}{*}{Low}} &
\multicolumn{1}{l?}{\multirow{2}{*}{High}} \\
\multicolumn{1}{|c||}{} & \multicolumn{1}{l|}{} & \multicolumn{1}{l|}{} & \multicolumn{1}{?l|}{} &
\multicolumn{1}{l?}{} \\
\cline{3-5} 
\multicolumn{1}{|c||}{} & \multicolumn{1}{l|}{} & \multicolumn{1}{l|}{\multirow{2}{*}{\%Walls}} & \multicolumn{1}{?l|}{\multirow{2}{*}{Low}} &
\multicolumn{1}{l?}{\multirow{2}{*}{High}} \\
\multicolumn{1}{|c||}{} & \multicolumn{1}{l|}{} & \multicolumn{1}{l|}{} & \multicolumn{1}{?l|}{} &
\multicolumn{1}{l?}{} \\
\cline{3-5} 
\multicolumn{1}{|c||}{} & \multicolumn{1}{l|}{} & \multicolumn{1}{l|}{\multirow{2}{*}{Topology}} & \multicolumn{1}{?l|}{\multirow{2}{*}{Low}} & 
\multicolumn{1}{l?}{\multirow{2}{*}{High}} \\
\multicolumn{1}{|c||}{} & \multicolumn{1}{l|}{} & 
\multicolumn{1}{l|}{} & \multicolumn{1}{?l|}{} & 
\multicolumn{1}{l?}{} \\
\hhline{|~||~|~|--} 
\hhline{|~||~|~|--} 
\hline
\multicolumn{1}{l}{} & & & &
\end{tabular}
\end{center}
\end{table}

A simplified simulation was created for accumulating a variety of observations in a given map. In a given run the observations were obtained by selecting an initial location within the empty occupancy grid at random and accumulating observations randomly in a tree-like fashion to a certain depth which was random as well. The simulation was run 5 times on each floorplan in each dataset. The collected observations in each dataset constituted the training set for the corresponding network.


\subsection{Evaluation}
\subsubsection{Evaluation Index}
A plausible characteristic of the presented mapping problem is the possibility of there being a substantial difference between the overall percentages of occupied and vacant spaces. Accordingly, the chosen performance indicator is the $F_1$ score, an accepted key index for testing the quality of a binary classifier over imbalanced data. This metric incorporates two measures with respect to the minority class -- in this case the {\it occupied/wall} class -- namely the precision and sensitivity/recall. The precision equals the percentage of actual walls out of the total of classified walls while the recall constitutes the percentage of correctly classified walls out of the total of actual walls. The $F_1$ score is the harmonic mean of the two. Following the probabilistic view, by which the {\it occupied/wall} class can be considered as {\it true} and the {\it vacant} class as {\it false},
\[
\begin{aligned}
    \text{precision} &= \frac{TP}{TP+FP}\\
    \text{recall} &= \frac{TP}{TP+FN}
\end{aligned}~~~,
\]
where $TP$ is the total of true positives, $FP$ is the total of false positives and $FN$ is the total of false negatives. The $F_1$ score can be expressed using these terms, yielding
\begin{equation}
    F_1 = 2 \cdot \frac{precision \cdot recall}{precision + recall} = \frac{2TP}{2TP+FN+FP}.
\label{eqn:f_score}
\end{equation}
Employing this evaluation index the map predictor's performance was gauged. 


\subsubsection{Test Results}
Two identically structured networks (as outlined in Section \ref{sec:predictor_architecture}) were trained separately on $\mathcal{D}_1$ and $\mathcal{D}_2$, one for each dataset. The performance of each network was measured across 30 floorplans from its corresponding dataset in terms of the 
$F_1$ scores. The baseline for comparison is the scores of the observations themselves.

\begin{figure}[tb]
  \includegraphics[width=\linewidth]{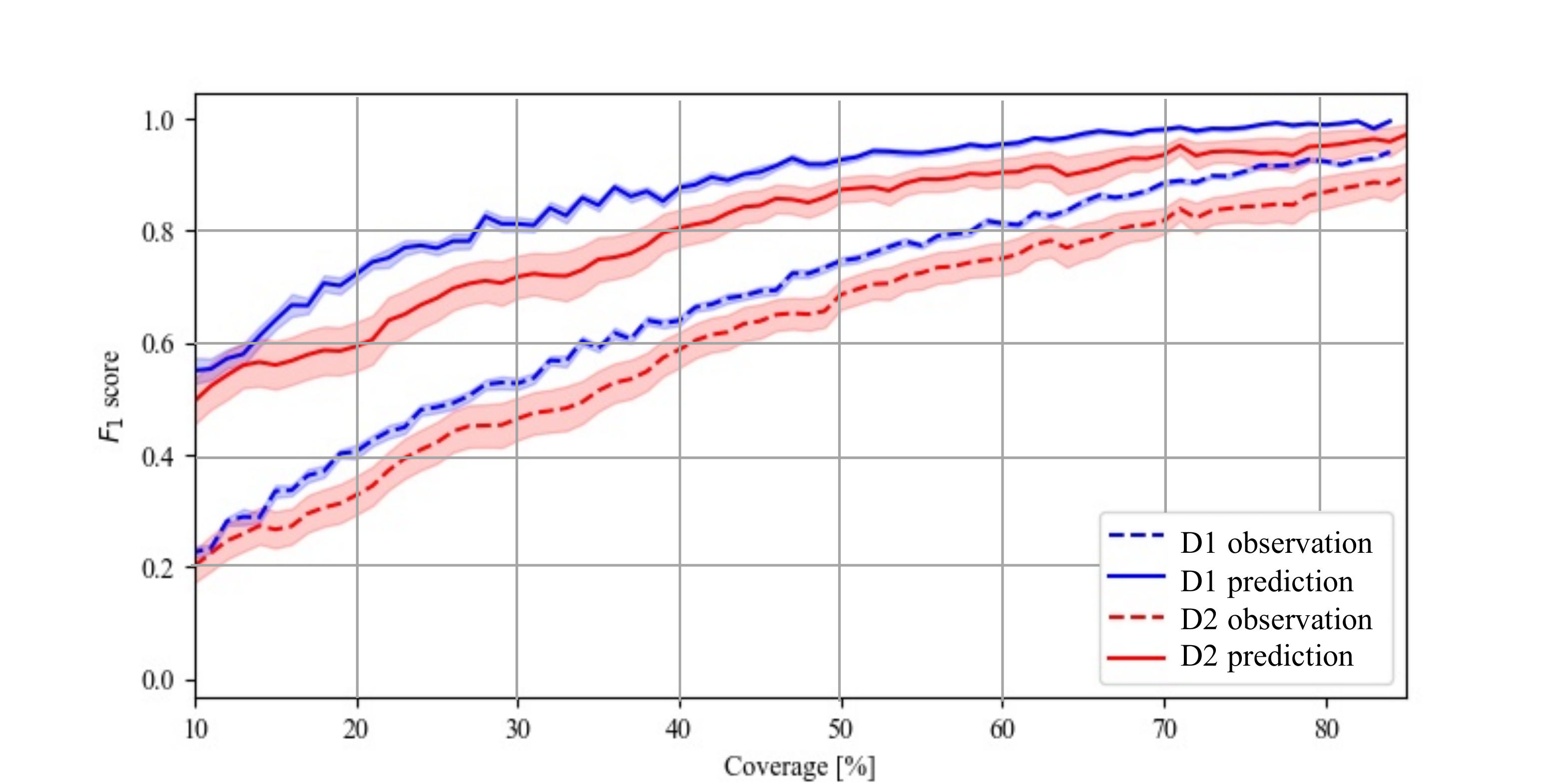}
  \caption{Prediction $F_1$ scores over the different datasets}
  \label{fig:f_scores}
\end{figure}

\figref{fig:f_scores} portrays the $F_1$ scores as a function of the number of observations. A significant improvement can be discerned in the performance obtained by the map predictor.

\section{Simulations}
To assess the potential increase in performance from integrating the map predictor several path planning algorithms were examined in the following two configurations: a) observation-based path planning and map construction, b) prediction-based path planning and map construction.

\subsection{Simulation Testbed}
A simplified grid world simulation was set up in Python, in which the mapping agent, situated in a certain cell, is free to move to any of its eight neighbouring cells, provided they are vacant. The agent is equipped with 16 fixed on-board range sensors arranged in equal angular intervals of 22.5$^{\circ}$ with an effective range of 20 pixels.

\subsection{Path Planning Algorithms}
With respect to path planning the occupancy grid is regarded as an undirected weighted graph, in which the nodes are the vacant cells and the vertices represent the contiguity of any two neighbouring nodes. The weight of each vertex is equal to the euclidean distance between the centers of its corresponding adjacent cells.

The partial observability of the graph which is to be explored eliminates the possibility of finding an optimal route. Therefore, heuristic-driven path planners, which strive to maximize some instantaneous utility, were examined \cite{yamauchi1997frontier,wirth2007exploration}. This utility, while not a direct representation of the original objective, is nonetheless a reflection of it, albeit in a more immediate sense.

During the mapping process, apart from its current location, the agent bases its decided path upon the accumulated observations, the map prediction, the constructed map, the visibility front, the map graph representation as well as some subset of the shortest distance paths in the graph.

Three path planning algorithms were considered, as detailed in the following.

\subsubsection{Random Exploration}
\textit{\textbf{Random Exploration}} follows the shortest path to a randomly chosen point on the visibility front.

\subsubsection{Greedy Selection}
Take a step along the shortest path to the point on the visibility front which yields the highest utility.
The utility of a point on the visibility front $v \in \mathcal{V}$ was chosen to be of the following form
\begin{equation}
    utility(v) = \frac{reward(v)}{1+cost(v)}~~~,
\end{equation}
where $reward(v)$ and $cost(v)$ are the estimated reward and cost to arrive at $v$, respectively.
Two variants (with respect to the definition of $utility(\cdot)$) of this algorithm were considered, as follows.

\textit{\textbf{Nearest Frontier}} chooses the point on the visibility front with the minimal shortest distance path from the agent's current position. i.e.,
\[
\begin{aligned}
reward(v) &\equiv 1, \\
cost(v) &\triangleq \text{shortest distance to point}~v.
\end{aligned}
\]

\textit{\textbf{Cost-Utility}}, by which the most cost effective point (with maximal benefit-to-effort ratio) on the visibility front is selected. In this version
\[
\begin{aligned}
reward(v) &\triangleq \text{potential confidence increase at point}~v, \\
cost(v) &\triangleq \text{shortest distance to point}~v.
\end{aligned}
\]

\subsection{Simulation Results \& Analysis}
\begin{figure}[t]
  \includegraphics[width=\linewidth]{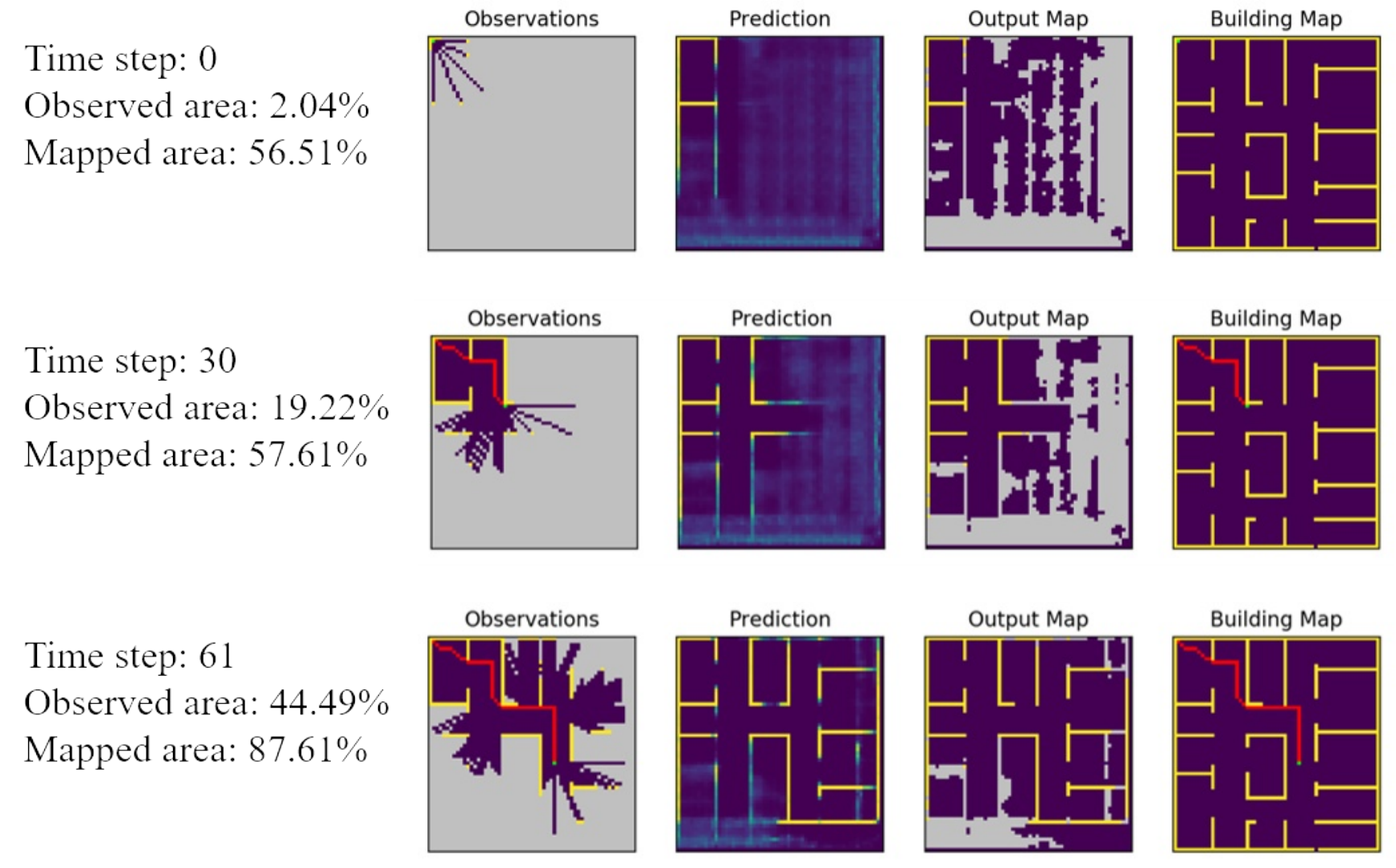}
  \caption{Example prediction-based path for {\it Cost-Utility} over a single floorplan from $\mathcal{D}_1$ for a required 85\% coverage at an 85\% level of confidence for both classification thresholds. The depicted outcome for each planner include the observations map (leftmost), the map prediction (second from the left), the constructed map (second from the right) and the ground truth map (rightmost). The observations and ground truth maps include the agent's location (green) and its traversed path (red). Vacant, occupied and unknown cells in each map are indicated in deep purple, yellow and grey, respectively}
  \label{fig:mapping_example}
\end{figure}

\begin{figure}[t]
\includegraphics[width=\linewidth]{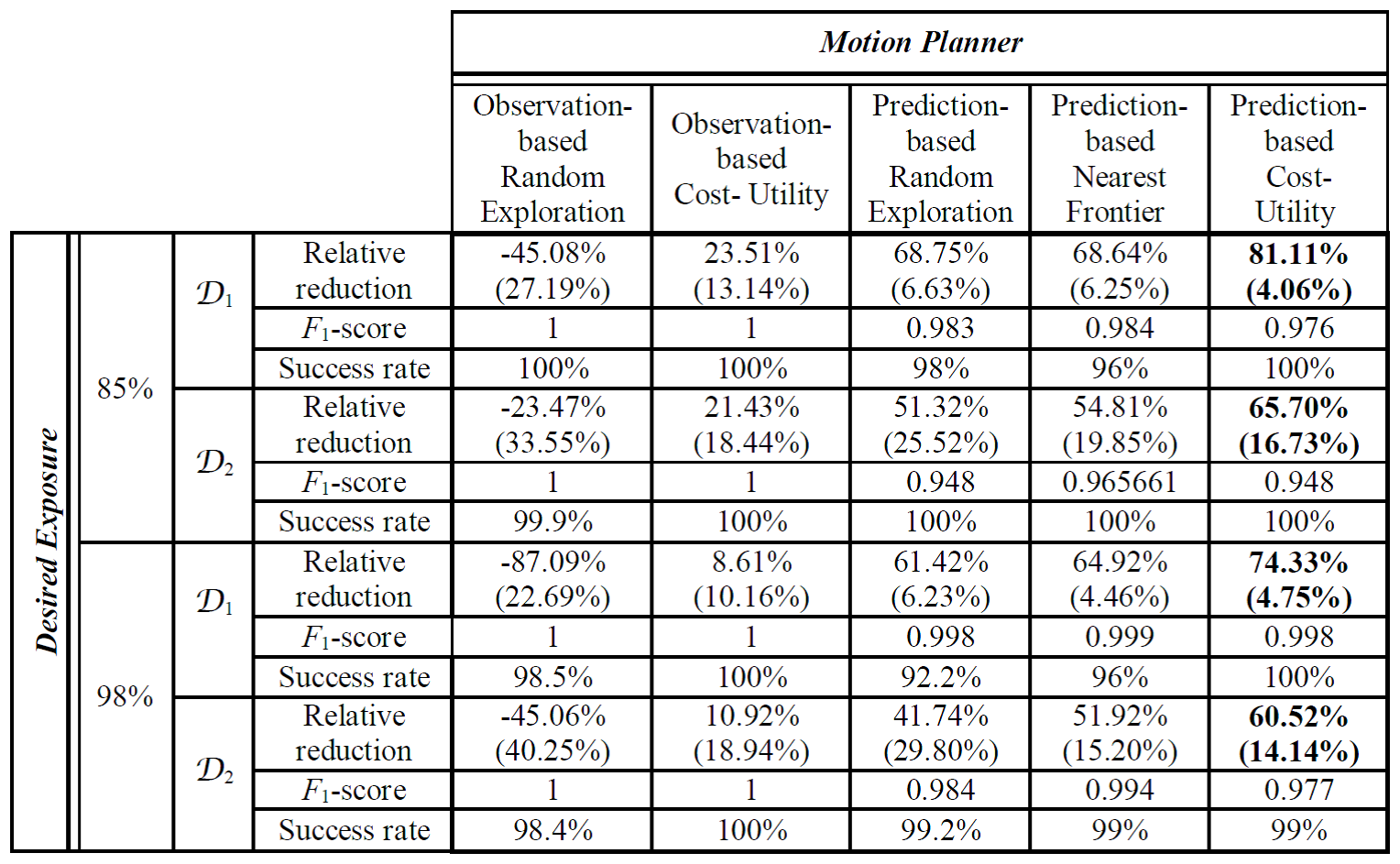}
  \caption{Mapping duration statistics of the different path planners in three prediction-based configurations over the different datasets. The results were obtained for a required coverage of 98\% and the classification confidence thresholds set to 95\% for occupancy and 93\% for vacancy}
  \label{fig:performance}
\end{figure}

The mapping performance is illustrated in \figref{fig:mapping_example} and summarized in \figref{fig:performance}. The results were obtained from simulation runs over 100 maps in each dataset where the agent's starting point was fixed to the top left interior cell. The evident advantage of integrating the map predictor in both the map construction and the path planning processes is discussed below.

An example run of the {\it Cost-Utility} path planner using prediction-based planning and map construction on a sample map from $\mathcal{D}_1$ is included in Fig. \ref{fig:mapping_example}. A comparison between the observations maps (leftmost column) and constructed maps (second column from the right) illustrates the substantial advantage of including the prediction in the map construction process. As demonstrated here, for marginal errors the map predictor holds the capacity to nearly double the map exposure for the final set of observations. 

\figref{fig:performance} includes statistical performance data, accumulated over 100 floorplans from each dataset. The performance is measured in terms of the relative mapping time reduction, with the observation-based nearest frontier path planner as the baseline. For the stochastic {\it Random Exploration} results from each map were averaged across 10 runs. The distinct advantage of using the map predictor is again evident, yielding a reduction of up to ~68\% for $\mathcal{D}_1$ and ~55\% for $\mathcal{D}_2$ on average in mapping time, as obtained for {\it Nearest Frontier}. For the superior {\it Cost-Utility} these numbers increase to ~80\% and ~65\%, respectively. Interestingly, even though the {\it Random Exploration} originally performed much worse compared to the baseline {\it Nearest Frontier}, it to managed to obtain substantial mapping time reductions in all cases when combined with the map predictor.
All prediction-based planners portray high $F_1$-scores and success rates in all cases. The situations in which a prediction-based planner fails is when the prediction seals off the agent, which can be easily avoided with the addition of a simple fail-safe logic.


\section{Future Work}
In future work the authors plan to extend the current problem to one in which aside from never having seen the environment that is to be explored the agent will have no prior knowledge as to the building's affiliation to a specific dataset. The initial approach will be to train a preliminary classifying network to determine the building's most probable denomination, according to which the appropriate map predictor will be employed.

\section{Conclusions}
The problem of autonomous indoor mapping in which the goal is to minimize the time to obtain a required percentage of coverage with a wanted level of certainty was considered. The proposed method by which to expedite the mapping process consisted of integrating a map predictor, realized as a pre-trained generative deep neural network, in both the path planning and the map construction.

Two distinct datasets of floorplans were examined, differing in several key aspects of the topology of the structures in each set. Two respective "specialized" map predictors were produced - each trained separately on a different dataset. The obtained map predictors were each examined in combination with several path planners: a simple random exploration and two variations of a greedy search, namely {\it nearest frontier} and {\it cost-utility} planners.

Simulations were run on representative groups of floorplans from each dataset. The proposed configuration of the fully integrated map predictor was compared to that of observation-based path planning and map construction. The results highlighted the significant potential improvement in terms of mapping time when incorporating the prediction in both the path planning and the map construction processes. Furthermore, a comparison between the different path planners revealed that without further insight simple random exploration can potentially at times outperform more sophisticated planners. Of the examined planners {\it Cost-Utility} greatly surpassed the others.






\bibliographystyle{elsarticle-num}
\bibliography{ml_based_planning}







\end{document}